\documentclass[10pt,twocolumn,letterpaper]{article}

\usepackage[pagenumbers]{iccv} 

\definecolor{iccvblue}{rgb}{0.21,0.49,0.74}
\usepackage[pagebackref,breaklinks,colorlinks,allcolors=iccvblue]{hyperref}

\definecolor{wdcolor}{RGB}{128, 0, 255}

\definecolor{wdqcolor}{RGB}{192, 0, 0}

\definecolor{qxlcolor}{RGB}{0, 0, 255}

\title{
TiMo: Spatiotemporal Foundation Model for Satellite Image Time Series
}

\author{%
  Xiaolei Qin\textsuperscript{1}$^*$,
  Di Wang\textsuperscript{1,2}$^*$, 
  Jing Zhang\textsuperscript{1}$^\dagger$,
  Fengxiang Wang\textsuperscript{3},
  Xin Su\textsuperscript{1},
  Bo Du\textsuperscript{1,2},
  Liangpei Zhang\textsuperscript{1}
  \\
  \thanks{$^*$: Equal contribution; $\dagger$: Corresponding author.}\\
\textsuperscript{1}Wuhan University
 \textsuperscript{2}Zhongguancun Academy
\textsuperscript{3}National University of Defense Technology\\
   {\centering}
    \texttt{\{qinxlei,d\_wang,xinsu.rs,dubo,zlp62\}@whu.edu.cn}\\
    \texttt{jingzhang.cv@gmail.com, wfx23@nudt.edu.cn}\\
    \\
}

\usepackage{multirow}

\begin{document}

\let\oldtwocolumn\twocolumn
\renewcommand\twocolumn[1][]{%
    \oldtwocolumn[{#1}{
\begin{center}
\centering
\includegraphics[width=1\linewidth]{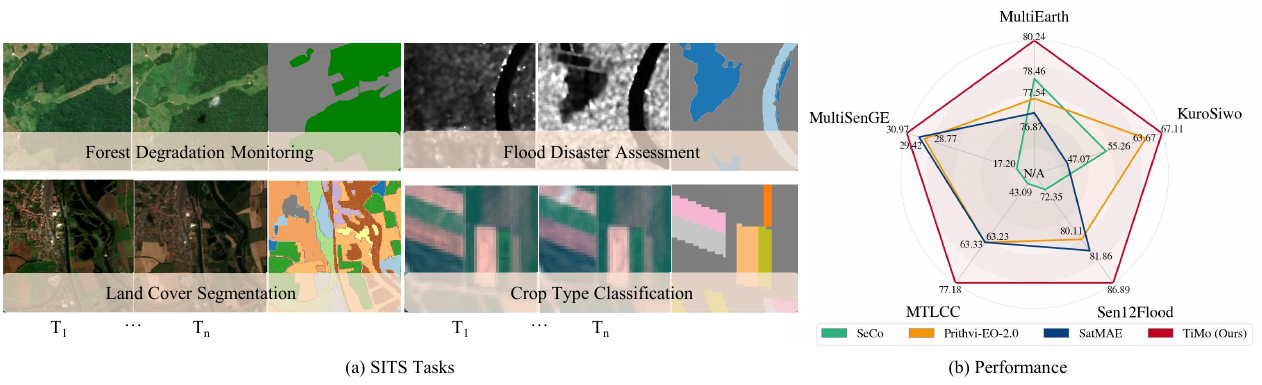}
\captionof{figure}{
TiMo surpasses existing spatiotemporal RSFMs, delivering superior performance across diverse SITS tasks, including forest monitoring, disaster assessment, ground-object recognition, and agricultural identification.
}
\label{fig:abstract}
\end{center}
    }]
}

\makeatletter
\def\thanks#1{\protected@xdef\@thanks{\@thanks
        \protect\footnotetext{#1}}}
\makeatother

\maketitle
\begin{abstract}
Satellite image time series (SITS) provide continuous observations of the Earth's surface, making them essential for applications such as environmental management and disaster assessment. However, existing spatiotemporal foundation models rely on plain vision transformers, which encode entire temporal sequences without explicitly capturing multiscale spatiotemporal relationships between land objects. This limitation hinders their effectiveness in downstream tasks. To overcome this challenge, we propose TiMo, a novel hierarchical vision transformer foundation model tailored for SITS analysis. At its core, we introduce a spatiotemporal gyroscope attention mechanism that dynamically captures evolving multiscale patterns across both time and space. For pre-training, we curate MillionST, a large-scale dataset of one million images from 100,000 geographic locations, each captured across 10 temporal phases over five years, encompassing diverse geospatial changes and seasonal variations. Leveraging this dataset, we adapt masked image modeling to pre-train TiMo, enabling it to effectively learn and encode generalizable spatiotemporal representations. Extensive experiments across multiple spatiotemporal tasks—including deforestation monitoring, land cover segmentation, crop type classification, and flood detection—demonstrate TiMo's superiority over state-of-the-art methods. Code, model, and dataset will be released at \href{https://github.com/MiliLab/TiMo}{TiMo}.

\end{abstract}    
\section{Introduction}
\label{sec:intro}

Recent advances in remote sensing and satellite missions have enabled higher-resolution satellite imagery with more frequent coverage, offering unprecedented opportunities to monitor Earth's surface \cite{tarasiou2023vits}. These repeated observations yield multi-temporal views, establishing satellite image time series (SITS) as vital tools for applications such as land cover monitoring, vegetation identification, and change detection \cite{wenger2022multisenge,li2024using,garnot2021panoptic,qin2025spatiotemporal}. Automated extraction of spatiotemporal patterns from SITS is essential to derive actionable insights and advance these critical domains.

Deep learning (DL) has become a cornerstone of satellite image time series (SITS) modeling due to its automated feature learning capabilities. However, conventional DL methods depend on large labeled datasets, which are frequently scarce \citep{cong2022satmae} due to annotation costs. Self-supervised learning (SSL) has gained traction as an efficient alternative, constructing supervisory signals from unlabeled data via pretext tasks that exploit inherent spatiotemporal structures \cite{he2020momentum,manas2021seasonal,ayush2021geography}. This paradigm minimizes dependence on manual annotations while maintaining strong downstream task performance \cite{cong2022satmae}.

Recently, SSL has driven the development of foundation models (FMs) in remote sensing, which learn universal representations from large-scale unlabeled data. FMs generalize across diverse downstream applications—even with limited labels—by pretraining on expansive datasets \cite{guo2024skysense}. For SITS, masked image modeling frameworks such as Masked Autoencoders (MAE) \cite{he2022masked,bao2022beit} and VideoMAE \cite{tong2022videomae} have been adapted \cite{cong2022satmae,szwarcman2024prithvi}, leveraging vision transformers (ViTs)~\cite{vit} to compute spatiotemporal self-attention across flattened token sequences. However, these models often overlook the inherent spatial alignment of SITS imagery and struggle to capture multiscale spatiotemporal patterns essential for land object analysis. On the other side, pretraining data also shapes FM performance. While datasets such as FMoW \cite{christie2018functional} and Sentinel-2 temporal composites \cite{manas2021seasonal} have been used for SITS pretraining \cite{cong2022satmae,ayush2021geography,manas2021seasonal}, their limited temporal cardinality (typically $\leq$5 timestamps) constrains the diversity of spatiotemporal change representation, reducing their effectiveness on downstream tasks.

To address these limitations, we propose TiMo, a novel hierarchical vision transformer foundation model for SITS analysis. TiMo introduces a spatiotemporal gyroscope attention (STGA) mechanism, which leverages the spatial alignment of SITS to capture correlations between tokens with identical temporal or spatial positions, where the term ``gyroscope'' refers to the geometric structure of attention regions. For pretraining, we curate MillionST, a million-scale dataset comprising 1 million Sentinel-2 images sampled from 100,000 geographic locations, each captured across 10 temporal phases over five years. This dataset encompasses diverse geospatial changes and seasonal variations. Inspired by \cite{ryali2023hiera}, we pretrain TiMo via spatiotemporal masked image modeling. To maximize the potential of FMs in capturing complex patterns within SITS \cite{bfm,hypersigma,oreole}, we scale TiMo to 300M parameters, enhancing its capacity to learn generalizable spatiotemporal representations. Extensive experiments across multiple spatiotemporal tasks, including deforestation monitoring, land cover segmentation, and flood detection, show that TiMo outperforms existing state-of-the-art methods, as shown in Figure \ref{fig:abstract}.

Our contributions are summarized as follows:

(1) We introduce TiMo, a novel spatiotemporal foundation model built on a hierarchical vision transformer tailored for unified SITS analysis. TiMo is pretrained via masked image modeling to learn multiscale spatiotemporal representations for diverse downstream tasks.

(2) We design a spatiotemporal gyroscope attention mechanism in TiMo, which leverages SITS' inherent spatial alignment to capture spatiotemporal relationships across satellite imagery.

(3) We curate MillionST, a large-scale pretraining dataset with 1 million samples from 100,000 geographic regions, each captured across 10 temporal phases over 5 years. This diversity enhances TiMo's ability to learn generalizable spatiotemporal features.

(4) Experiments demonstrate TiMo's superiority over existing SITS foundation models across tasks like land-cover classification, disaster assessment, and agricultural mapping. Additionally, TiMo exhibits scalability, sample efficiency, and interpretability, further validating its utility.

\section{Related work}
\subsection{Remote Sensing Foundation Models}

Inspired by the success of FMs in the natural image domain \cite{swin_v2,v_moe,vit_g,xu2021vitae,internimage}, researchers have increasingly explored the development of remote sensing FMs (RSFMs), with vision transformers emerging as the predominant model architecture. For instance, some studies have leveraged large-scale labeled dataset to pre-train RSFMs \citep{bastani2023satlaspretrain, wang2022empirical} using advanced vision transformer architectures \cite{swin, zhang2023vitaev2}. In addition, domain-specific features, such as spectral information \cite{spectralgpt,hypersigma,usat}, geographic locations \cite{csp, geoclip, satclip}, geometric and physical attributes of land objects \cite{wang2022advancing,sun2022ringmo,matter,Cross-ScaleMAE,MA3E}, and modal properties \cite{msGFM,dofa,senpamae,omnisat,anysat}, have been extensively explored to refine pre-training strategies and model design \cite{reed2023scale, mendieta2023towards, wang2024mtp, cspt, tov, cmid}. However, these pioneering RSFMs mainly focused on learning generalizable feature representations by pre-training on single-temporal RS images. As a result, they are better suited for time-independent downstream tasks such as scene classification, semantic segmentation, and object detection.

\begin{figure*}[t] 
	\includegraphics[width=\linewidth]{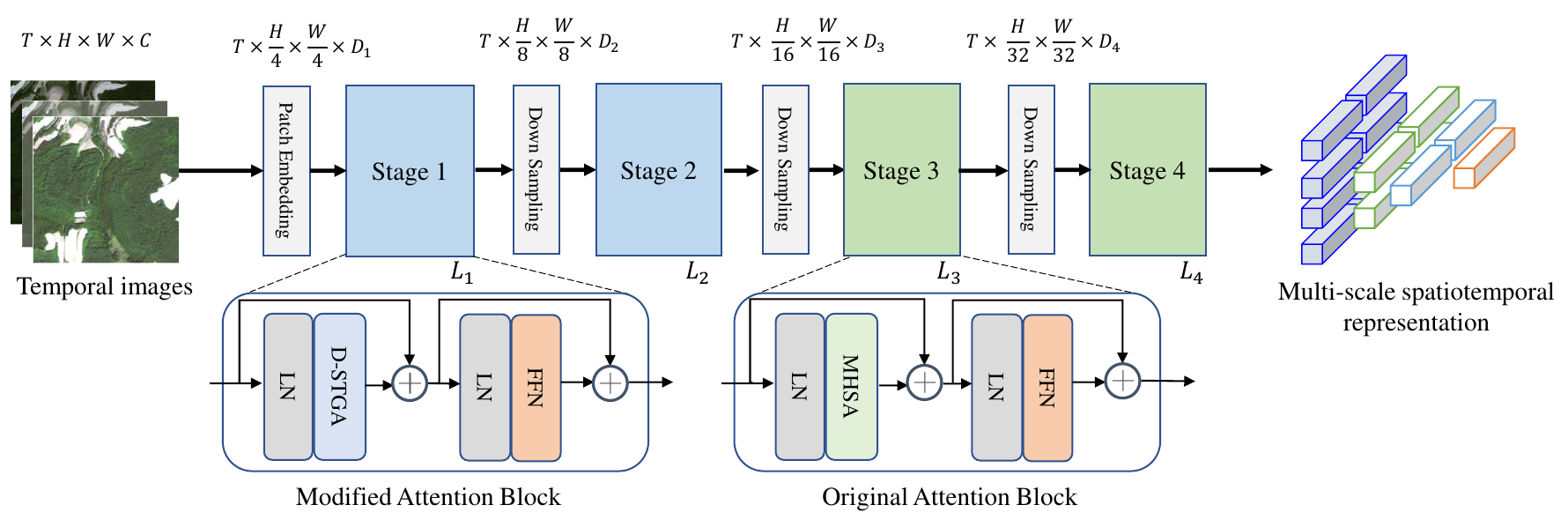}
	\caption{
    \textbf{Overview of TiMo's Architecture}: TiMo follows a four-stage hierarchical design, where each stage comprises attention layers and feed-forward networks following a downsampling layer. To enhance spatio-temporal representation learning, TiMo replaces Multi-head Self-Attention (MHSA) with a novel and efficient Differential Spatiotemporal Gyroscope Attention (D-STGA) in the first two stages.
    }
	\label{Finetuning}
\end{figure*}

\subsection{SITS Foundation Models}

Unlike those FMs that focus solely on single-temporal imagery, recent studies have begun incorporating temporal information from revisited satellite images \citep{manas2021seasonal,mall2023change,ayush2021geography,cong2022satmae,a2mae,guo2024skysense,szwarcman2024prithvi,wang2023ssl4eo,nakayama2024spatio}. For example, some approaches leveraged contrastive learning (CL) by constructing positive and negative pairs from temporal remote sensing images~\cite{manas2021seasonal,mall2023change,ayush2021geography}. To simplify the complex process of generating negative pairs, SkySense \cite{guo2024skysense} employs a teacher-student framework to apply CL to augmented temporal inputs. While these CL-based FMs rely heavily on data augmentation \cite{he2022masked}, alternative approaches \citep{cong2022satmae,szwarcman2024prithvi} adopt masked image modeling \cite{bao2022beit} for pre-training. These SITS FMs typically extend the MAE \cite{he2022masked} and plain ViT \cite{vit} architectures to encode patches across both spatial and temporal dimensions. However, they often overlook the underlying spatiotemporal relationships within SITS data. In contrast, we introduce TiMo, a novel hierarchical vision transformer designed for multiscale spatiotemporal representation learning. TiMo builds upon the STGA module, which simultaneously captures contextual relationships within the same spatial or temporal dimensions across all timestamps. Given the progressive downsampling nature of its hierarchical structure, we adopt the self-supervised learning paradigm of Hiera \cite{ryali2023hiera} and apply masked image modeling to multi-temporal images for pre-training.

\subsection{Pre-training Data for SITS Foundation Models}

Satellite revisit cycles generate vast amounts of global-scale SITS, significantly enhancing the SSL pre-training of spatiotemporal FMs. The Sentinel-2 mission\footnote{https://sentiwiki.copernicus.eu/web/s2-mission}, with its twin satellites revisiting the same location every five days, provides high-temporal-resolution multispectral imagery. SeCo \cite{manas2021seasonal} curated Sentinel-2 SITS from circular regions with a 100 km radius across 10,000 cities worldwide, with each location contributing five images representing different seasons over a year. CACo \cite{mall2023change} further refined this sampling approach by reducing spatial distances relative to city centers. The widely recognized FMoW dataset \cite{christie2018functional} has played a crucial role in advancing RSFMs for SITS \cite{cong2022satmae, ayush2021geography}. It consists of high-resolution satellite image time series designed to classify images into 62 distinct categories. However, many existing pre-training datasets suffer from limited temporal coverage. For instance, approximately 70\% of FMoW samples contain time series with five or fewer timestamps \cite{ayush2021geography}, which may constrain the effectiveness of pre-trained RSFMs in downstream tasks. To address these limitations, this paper introduces MillionST, a large-scale SITS dataset specifically designed for spatiotemporal pre-training. Collected from 100,000 geographic locations, MillionST comprises approximately one million samples, with each location contributing images from 10 timestamps spanning five years. This extensive spatiotemporal diversity enhances RSFMs pre-training, \eg, our TiMo series.

\section{Method}

\subsection{Overall Architecture of TiMo}
To more effectively capture the spatiotemporal variations—such as differing scales and shapes—inherent in SITS data, we employ a hierarchical vision transformer design in our TiMo model, which is capable of generating multiscale pyramid features. As shown in Figure \ref{Finetuning}, TiMo consists of four stages separated by downsampling layers. In the early stages, we replace the standard multi-head self-attention (MHSA) with a novel Spatiotemporal Group Attention to enhance the extraction of spatiotemporal features. However, in the deeper stages, where progressive downsampling results in more abstract, low-resolution features, the STGA module may be less effective, so we retain the original attention mechanism. The optimal configuration for each stage will be validated via empirical study.

\textbf{Patch Embedding and Downsampling Layers} 
Given a group of input SITS with dimensions $T\times H\times W \times C$, where $T$ represents the temporal length and $H$,$W$, and $C$ denote the height, width, and number of channels, respectively, we begin by applying patch embedding to each image across different timestamps. The patch embedding layer consists of two convolutional operations: a $7 \times 7$ convolution with a stride of 2 and padding of 3, followed by a $2 \times 2$ convolution with a stride of 2. As a result, each image is transformed into $\frac{H}{4} \times \frac{W}{4}$ tokens with a feature dimension of $D_1$. For simplicity, we obtain the spatiotemporal positional encoding of each token by concatenating its temporal and spatial positional encodings—both represented as 1D vectors—along the channel dimension. The temporal positional embedding follows the approach introduced by \cite{vaswani2017attention}, while the spatial positional encoding is adopted from \cite{he2022masked}. To facilitate hierarchical processing, the model consists of four stages connected by downsampling layers. Each downsampling layer applies a $3 \times 3$ convolution with a stride of 2 and padding of 1, ensuring a downsampling ratio of 2 between consecutive stages. Additionally, the output feature maps of each downsampling layer have an expanded channel dimension, doubling in size at each stage.

\textbf{Transformer Blocks}
TiMo employs two types of attention blocks: one with the proposed attention module and the other using the original MHSA. Aside from the attention mechanism, both blocks share the same structure, consisting of a layer normalization (LN) layer, a feed-forward network (FFN), and residual connections. The details of these components follow that of the Swin Transformer \cite{swin}.

\textbf{Multi-head Self-Attention}
MHSA is a fundamental component of vision transformers, comprising several parallel self-attention mechanisms—each commonly referred to as a head. It is defined by:
\begin{equation}
SA(\textbf{F})=softmax(\frac{\textbf{Q}\textbf{K}^\top}{\sqrt{D'}})\textbf{V},
\end{equation}
where $\textbf{Q}\in\mathbb R^{N\times D'}$,  $\textbf{K}\in\mathbb R^{N\times D'}$,  $\textbf{V}\in\mathbb R^{N\times D'}$ represent the query,
key, and value matrices, respectively, which are derived from $\textbf{F}\in\mathbb R^{N\times D}$ by three linear layers. Here, $N$ denotes the total number of tokens across all temporal dimensions, and $D$ is the channel dimension of the tokens. The head dimension $D'$ is defined such that $D=hD'$, with $h$ being the number of heads. Specifically, for the $i$th stage, $N$ is calculated as: $N = \frac{H}{2^{i+1}}\cdot \frac{W}{2^{i+1}} \cdot T$, $i\in \{1,2,3,4\}$. In the TiMo architecture, $\textbf{F}$ is the feature output from the first LN layer in MHSA blocks. In subsequent sections, we omit different heads for convenience.

\begin{table}[t]
\setlength{\tabcolsep}{2pt}
\renewcommand{\arraystretch}{1.1}
\caption{Configuration for TiMo variants. $S_i$ denotes the $i$th stage.}
\resizebox{\linewidth}{!}
{
\begin{tabular}{l|cccc|cccc|cccc}
\hline
\multirow{2}{*}{Version} & \multicolumn{4}{c|}{Dims ($D$)} & \multicolumn{4}{c|}{Heads $(h)$}& \multicolumn{4}{c}{Depths $(L)$} \\ \cline{2-13} 
                         & $S_1$   & $S_2$  & $S_3$   & $S_4$   & $S_1$   & $S_2$   & $S_3$   & $S_4$   & 
                         $S_1$   & $S_2$   & $S_3$   & $S_4$  \\ \hline
TiMo-Base                & 128  & 256 & 512  & 1024 & 4    & 8    & 16   & 32 & 2  & 2  & 18   & 2   \\
TiMo-Large               & 384  & 768 & 960  & 1536 & 6    & 12   & 24   & 48 & 2  & 2  & 18   & 2 \\
TiMo-Huge              & 512  & 1024 & 1280 & 2048 & 8    & 16   & 32   & 64 & 3  & 3  & 22   & 3 \\ 
\hline
\end{tabular}
}
\label{tab:arch}
\end{table}

\textbf{Parameter Scaling} Following \cite{swin}, we design a series of TiMo with various sizes, \ie, TiMo-Base, TiMo-Large and TiMo-Huge. 
These versions differ in the dimensions of features, number of heads, and the depths in each stage, as shown in Table \ref{tab:arch}. Other configurations between these versions are always the same. For example, the MLP expansion ratio is set to 4.

\subsection{STGA: Spatiotemporal Gyroscope Attention}
\label{subsec:stga}

\begin{figure}
	\centering
	\includegraphics[width=0.9\linewidth]{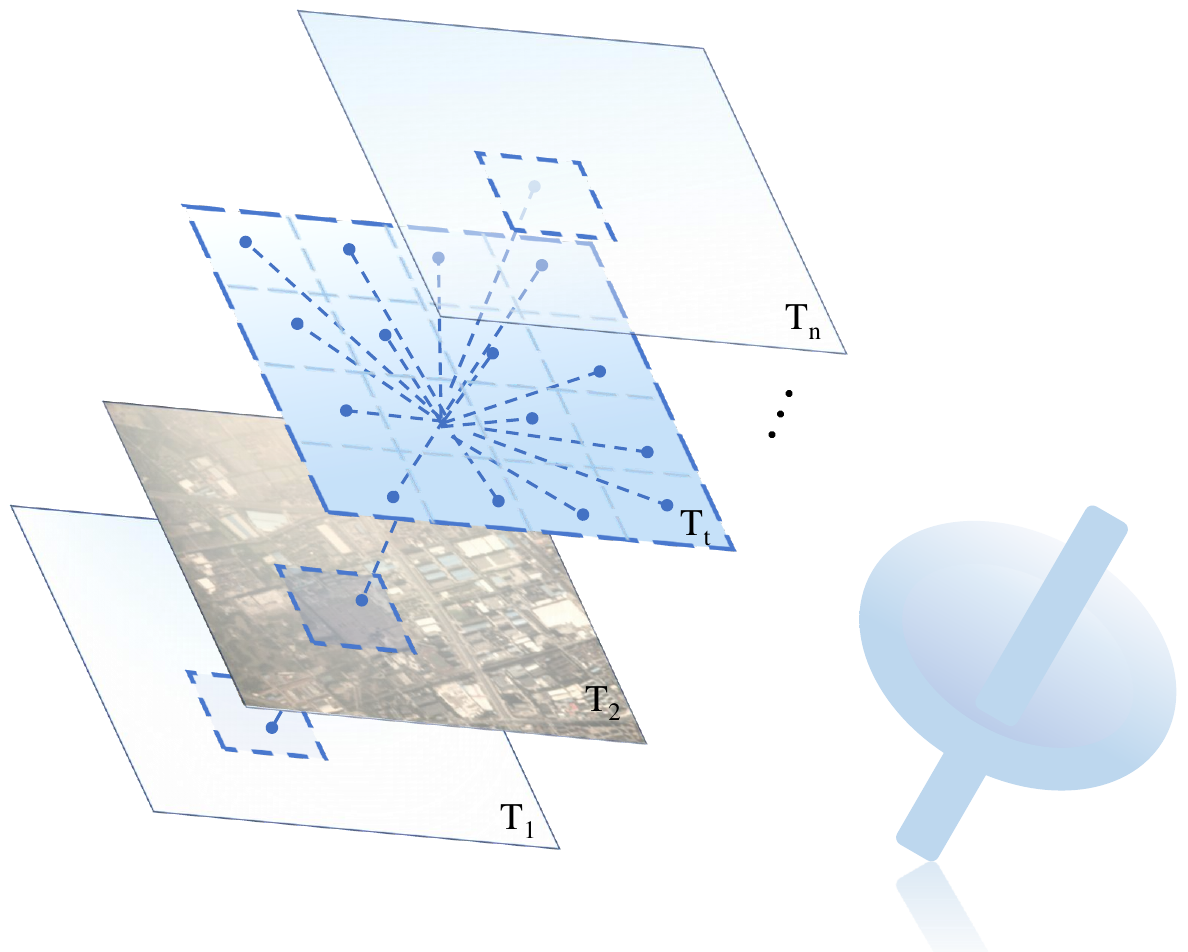}
	\caption{Illustration of Spatiotemporal Gyroscope Attention.}
	\label{fig:STGA}
\end{figure}

In this section, we introduce the STGA mechanism. Rather than calculating similarities between all tokens as in traditional MHSA used in standard vision transformers, STGA computes correlations only among tokens that share the same spatial or temporal positions, thereby enhancing the model's ability to capture spatiotemporal dynamics.

Let $\textbf{Q}\in\mathbb R^{T \times N_p\times N_p\times D'}$, $\textbf{K}\in\mathbb{R}^{T \times N_p \times N_p \times D'}$, and $\textbf{V}\in\mathbb{R}^{T \times N_p \times N_p \times D'}$, where $N_p = \sqrt{N/T}$ denotes the number of tokens along the height or width of an image (assuming $H=W$). Consider a query vector $q_p \in \mathbb{R}^{1 \times D'}$ from $\textbf{Q}$ at position $p(x_0, y_0, z_0)$, where $x$ and $y$ lie in the plane $\mathcal{P}$ and $z$ represents the time axis $\mathcal{T}$. Define the set of positions $\textbf{u} = \{(x, y, z) \mid z = z_0 \lor (x = x_0 \land y = y_0)\}$ within the three-dimensional spatiotemporal space $\mathcal{S} = \mathcal{P} \times \mathcal{T}$ (with $\times$ denoting the Cartesian product), ensuring that each element in $\textbf{u}$ shares either the same temporal or spatial coordinate as $p$. From $\textbf{K}$, we then extract a vector set $\mathbf{K}_\textbf{u} \in \mathbb{R}^{(T + N_p \times N_p - 1) \times D'}$ corresponding to these positions. Finally, the attention scores for each spatiotemporal location are computed as follows:
\begin{equation}
\textbf{A}_p=q_p\mathbf K^{\top}_\textbf{u} \label{au},
\end{equation}

where $\textbf{A}_p \in \textbf{A}$ denotes the affinity between $q_p$ and $\textbf{K}_\textbf{u}$. Similarly, the overall attention matrix is given by $\textbf{A} \in \mathbb R^{(T\times N_p \times N_p) \times (T+N_p\times N_p-1)}$. We then apply a softmax operation along the last dimension of $\textbf{A}$ to compute the attention map $\textbf{A}'$, where each $\textbf{A}'_p \in \mathbb R^{1 \times (T+N_p\times N_p-1)}$ determines the contribution of each location to the representation at position $p$. 

Following a similar procedure, we extract a set of vectors $\mathbf{V}_\textbf{u} \in \mathbb R^{(T+N_p\times N_p-1) \times D'}$ from $\textbf{V}$. The output for position $p$, denoted as $\textbf{O}_p \in \mathbb{R}^{1\times D'}$, is then computed according to:
\begin{equation}
\textbf{O}_p=\textbf{A}'_p\mathbf{V}_\textbf{u} \label{fp1}.
\end{equation}

The final feature $\textbf{F}_p\in \mathbb R^{1 \times D}$ is obtained by concatenating the outputs $\textbf{O}_p$ from different heads and passing the result through a linear layer. Consequently, for all spatiotemporal positions, we obtain $\textbf{F} \in \mathbb R^{T \times N_p \times N_p \times D}$.

This process reveals that the shape of the core variable $\textbf{u}$ resembles an extreme gyroscopic form in the spatiotemporal domain, which is the inspiration behind the name ``STGA'', as illustrated in Fig.~\ref{fig:STGA}.

\subsection{D-STGA: Differential Spatiotemporal Gyroscope Attention}
\begin{figure}[!h]
	\includegraphics[width=1\linewidth]{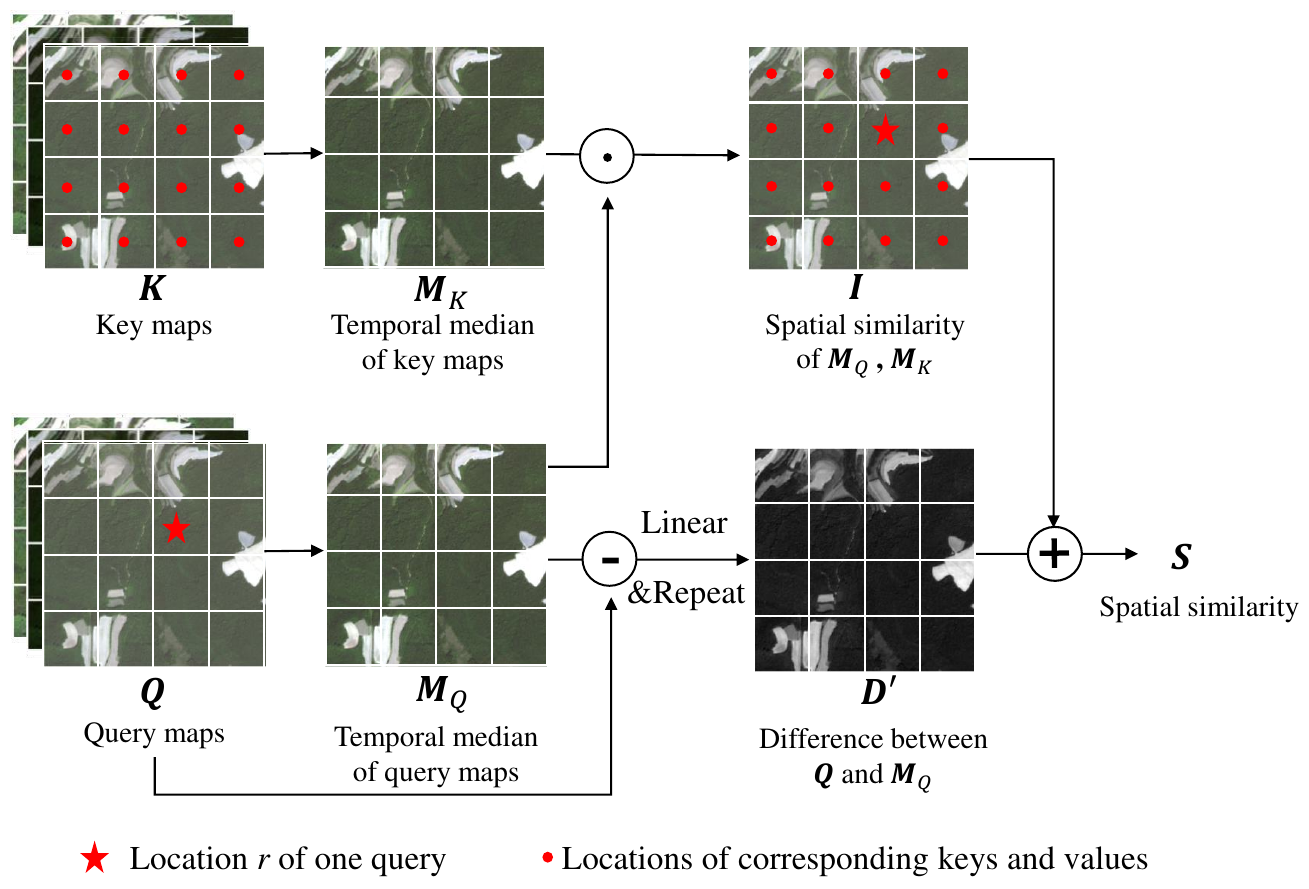}
	\caption{Diagram of calculating spatial similarity in D-STGA.}
	\label{DGA}
\end{figure}

Although STGA can model spatiotemporal variations, its computational cost increases rapidly with the temporal length \(T\), particularly when computing the attention matrix. With a fixed input size, the theoretical computational complexity of obtaining \(\textbf{A}\) in STGA is $O((T\times N_p \times N_p) \times (T+N_p\times N_p-1))\approx O(T^2)$, which motivates us to develop a more lightweight implementation.

We observe that multi-temporal observations contain both variant and invariant regions; that is, the spatial status at any time can be inferred from temporal relationships. This propagation can be approximated by using temporal differences, leading to a novel differential spatiotemporal gyroscope attention (D-STGA).

Given $\textbf{Q}\in\mathbb R^{T \times N_p\times N_p\times D'}$, $\textbf{K}\in\mathbb R^{T \times N_p\times N_p\times D'}$, and $\textbf{V}\in\mathbb R^{T \times N_p\times N_p\times D'}$, we first extract the median temporal features from \(\textbf{Q}\) and \(\textbf{K}\) (using the \texttt{pytorch.median} function) to mitigate the effect of abnormal observations (\eg, clouds in SITS). This yields \(\textbf{M}_Q \in \mathbb{R}^{N_p\times N_p\times D'}\) and \(\textbf{M}_K \in \mathbb{R}^{N_p\times N_p\times D'}\). We then compute their similarity:
\begin{equation}
\textbf{I}=\textbf{M}_Q\textbf{M}_K^{\top} \label{S}.
\end{equation}

Here, \(\textbf{I}\) represents the temporal-invariant spatial similarity. To recover spatial information for all timestamps, we compute the difference sequence \(\{\textbf{D}_t \mid t=1,2,\dots,T\}\), where $\textbf{D}_t = \textbf{Q}_t - \textbf{M}_Q$,
with \(\textbf{Q}_t \in \mathbb{R}^{N_p\times N_p\times D'}\) denoting the query at time \(t\). Next, we apply a linear projection (comprising a Linear layer, BatchNorm, and ReLU) to reduce the channel dimension of \(\textbf{D}_t\) to 1. Repeating the result \(N_p \times N_p\) times produces \(\textbf{D}'_t \in \mathbb{R}^{(N_p \times N_p) \times (N_p \times N_p)}\), which captures the spatial discrepancy.

The spatial status for each timestamp is then obtained by adding \(\textbf{D}'_t\) to \(\textbf{I}\): $\textbf{S}_{:,t} = \textbf{D}'_t + \textbf{I},\quad t=1,2,\dots,T.$
Thus, the complete spatial status set is $\textbf{S} = \{\textbf{S}_{:,1},\dots,\textbf{S}_{:,T}\} \in \mathbb{R}^{T \times (N_p \times N_p) \times (N_p \times N_p)}.$
In this way, it efficiently calculates the spatial similarity across all timestamps. The above process is illustrated in Figure \ref{DGA}.

In addition to spatial status, temporal relationships must also be estimated. For a spatial location \(r\) at time \(t\), we compute the temporal correlation by multiplying its query token \(\textbf{Q}_{r,t} \in \mathbb{R}^{1 \times D'}\) with the key tokens \(\mathbf{K}_\textbf{o} \in \mathbb{R}^{(T-1) \times D'}\) from the same spatial position at other timestamps:

\begin{equation}
\textbf{T}_{r,t}=\textbf{Q}_{r,t}\mathbf{K}^\top_\textbf{o} \label{tp},
\end{equation}

where \(\textbf{T}_{r,t} \in \mathbb{R}^{1 \times (T-1)}\). We then concatenate the spatial affinity map \(\textbf{S}_{r,t} \in \mathbb{R}^{1 \times (N_p \times N_p)}\) with the temporal correlation \(\textbf{T}_{r,t}\) to obtain the spatiotemporal similarity map \(\textbf{A}_{r,t} \in \mathbb{R}^{1 \times (T+N_p \times N_p -1)}\). The attention score is then derived as described in Sec.~\ref{subsec:stga}, with subsequent steps—such as applying the softmax, matrix multiplication, and linear projection—following the same procedure as in STGA.

Compared to STGA, D-STGA circumvents the vector inner product, reducing the computational complexity of obtaining \(\textbf{A}\) to \(O(T)\).

\subsection{Pre-training and Fine-tuning}

\textbf{Pre-training} Masked image modeling methods, such as MAE \cite{he2022masked}, are well-regarded for their performance and efficiency and have been widely adopted in pre-training remote sensing and SITS foundation models \cite{wang2022advancing,reed2023scale,cong2022satmae,MA3E,Cross-ScaleMAE} that use the vanilla ViT \cite{vit}. However, TiMo employs a hierarchical vision transformer network with convolutional downsampling layers, which prevents the direct application of MAE for pre-training. To address this, we follow Hiera \cite{ryali2023hiera} to adopt an MAE-like approach specifically designed for hierarchical architectures. Although it was originally used to pre-train MViTv2 \cite{MVITV2} on natural images and video, we have successfully applied it to our architecture without modifying the attention mechanism (\ie, we use MHSA in all blocks during pre-training). Notably, while Hiera uses a mask unit of 2$\times$32$\times$32 pixels for video data to reduce redundancy between frames, SITS data are sparser with longer temporal intervals; hence, we adopt a mask unit of 1$\times$32$\times$32 pixels. \textit{In this way, we successfully pre-trained TiMo, a hierarchical vision transformer foundation model for SITS.} Additional details on pre-training are provided in the Supplementary Material.

\textbf{Fine-tuning} During fine-tuning on downstream tasks, we reuse the attention layer weights as in \cite{wang2022advancing, vitdet} and replace the standard MHSA in the early stages with the proposed STGA (or D-STGA) to better model spatiotemporal contexts. For the parameters introduced by the linear projection used to compute the spatial discrepancy in D-STGA—which lack pre-trained weights—we initialize them randomly. The output features from all four stages of TiMo are used for segmentation tasks, while the feature from the final stage is used for classification.

\section{Experiments}
\subsection{Pre-training Dataset: MillionST}
\begin{figure}
	\centering
	\includegraphics[width=0.8\linewidth]{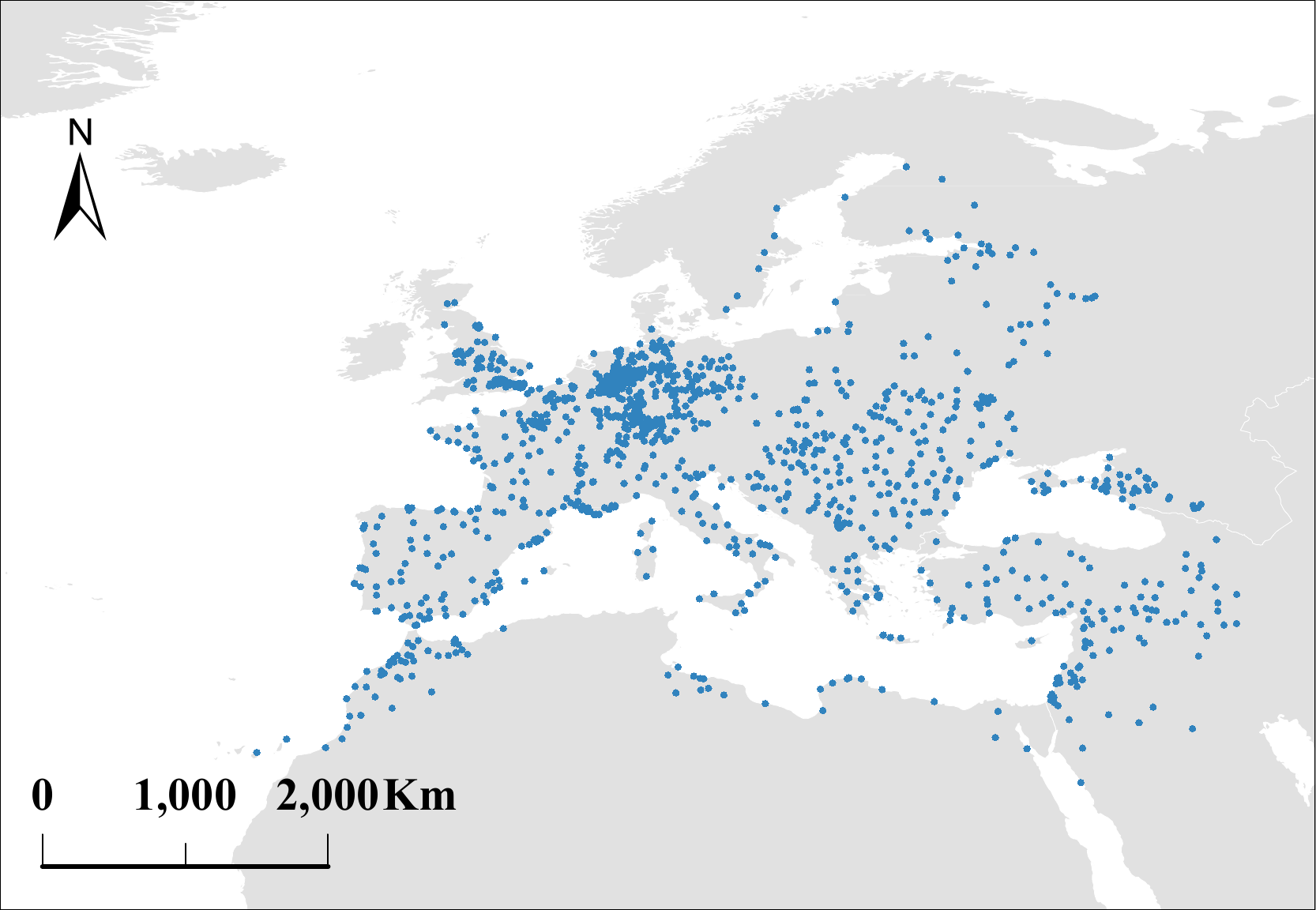}
	\caption{City distribution for MillionST data sampling.}
	\label{fig:millionst_city}
\end{figure}

We collected Sentinel-2 multi-spectral SITS data from 100,000 locations spanning 1,317 cities—primarily in Europe, North Africa, and West Asia—all of which rank among the 10,000 most populous cities globally, as shown in Fig.~\ref{fig:millionst_city}. Following the approach of \cite{manas2021seasonal}, coordinates were sampled from a Gaussian distribution centered on each city with a 50 km standard deviation. For every location, we compiled a temporal sequence of images starting on June 30, 2017, with captures occurring at six-month intervals. Consequently, the dataset comprises SITS data from ten distinct timestamps covering the period from 2017 to 2022. Additional details regarding the sampling cities are provided in the Supplementary Material.

\subsection{Deforestation Monitoring} 
The MultiEarth dataset \cite{cha2023multiearth} is a multi-modal RS dataset designed to monitor deforestation in the Amazon rainforest. It utilizes Sentinel-2 RGB temporal images, with the training split comprising 1,565 images that include pixel-level annotations distinguishing forest from deforested areas. The dataset is divided into training, validation, and testing sets in a 1:1:8 ratio. Each image is 256 $\times$ 256 pixels in size, and three timestamps are randomly selected for both training and testing. Model performance is evaluated using the mean intersection over union (mIoU) metric. For further details on the implementation, as well as those in the following sections, please refer to the Supplementary Material.

We compare TiMo against recent SOTA multi-temporal RSFMs, including SeCo \cite{manas2021seasonal}, CACo \cite{mall2023change}, GASSL \cite{ayush2021geography}, Prithvi-EO-2.0 \cite{szwarcman2024prithvi}, and SatMAE \cite{cong2022satmae}. To assess the impact of the established pre-training dataset, we also pre-trained SatMAE on our MillionST, which is denoted as SATMAE\dag. The quantitative results in Table \ref{deforestTable} indicate that: (1) MillionST is more effective for pre-training spatiotemporal foundation models—likely due to its richer set of timestamps compared to the FMoW dataset \cite{christie2018functional} originally used for SatMAE, and (2) TiMo-Base outperforms the competing methods, even with relatively fewer parameters than Prithvi-EO-2.0 and SatMAE. Moreover, TiMo's performance scales favorably with increased model size.

\begin{table}
\small
\centering
\caption{Comparison of different models on MultiEarth dataset, \dag: pre-training on MillionST.}
\resizebox{\linewidth}{!}{
\begin{tabular}{llcc} 
\hline
Method  &Backbone&\#Parameter&  mIoU \\ 
        \hline
          SeCo\cite{manas2021seasonal}&ResNet-50&26M&0.7846 \\
          CACo\cite{mall2023change}&ResNet-50&26M&0.7692 \\
          GASSL\cite{ayush2021geography}&ResNet-50&26M&0.7638\\
          Prithvi-EO-2.0\cite{szwarcman2024prithvi}&ViT-Large&300M&0.7754\\
          SatMAE\cite{cong2022satmae}&ViT-Large&307M&0.7687\\
          SatMAE\cite{cong2022satmae}\dag&ViT-Large&307M&0.7828 \\
          \hline
         TiMo-Base &TiMo-Base&91M&0.7860\\
        TiMo-Large &TiMo-Large&298M&\textbf{0.8024}\\
        
\hline\label{deforestTable}
\end{tabular}
}
\end{table}

\subsection{Multi-class Land Cover Segmentation}

\label{multisenge:intro}

We perform multi-class land cover segmentation using the MultiSenGE dataset~\cite{wenger2022multisenge}, which generates a single land cover map from multi-temporal images. The dataset comprises 8,157 time series samples collected from 14 distinct Sentinel-2 tiles, each with 10 bands and a temporal length of 12. Each image has a spatial resolution of 256 $\times$ 256. For our experiments, we allocate 8 tiles for training, 3 for validation, and the remaining 3 for testing. In all experiments, we randomly select 3 temporal data points from each image time series to train all comparison methods.

The experimental results for multi-class land cover segmentation are presented in Table~\ref{tab:multisenge}. As shown, SeCo, CACo, and GASSL achieve relatively lower accuracy. Although both Prithvi-EO-2.0 and SATMAE perform well, our proposed TiMo-Base and TiMo-Large models outperform them, underscoring the robust capabilities of our approach. Moreover, as the model's parameter count increases, its performance continues to improve.

\begin{table}
\small
    \centering
    \caption{Comparison of different models on MultiSenGE dataset.}\setlength{\tabcolsep}{8pt}
    \label{tab:multisenge}
    \begin{tabular}{llcc}
    \hline
        Method  &Backbone&\#Parameter& mIoU   \\ 
        \hline
          SeCo\cite{manas2021seasonal}&ResNet-50&26M &0.1720 \\
          CACo\cite{mall2023change}&ResNet-50&26M&0.1789 \\
          GASSL\cite{ayush2021geography}&ResNet-50&26M&0.1712\\
          Prithvi-EO-2.0\cite{szwarcman2024prithvi}&ViT-Large&300M&0.2877\\
          SatMAE\cite{cong2022satmae} &ViT-Large&307M&0.2942\\
          \hline
         TiMo-Base&TiMo-Base&91M&0.2977\\
         TiMo-Large&TiMo-Large&298M&\textbf{0.3097}\\
         
         \hline
    \end{tabular}
\end{table}

\subsection{Crop Type Classification}

To evaluate the models' ability to handle longer time series, we conducted an experiment using the MTLCC dataset \cite{russwurm2018multi}, which is designed for crop type classification. This dataset comprises Sentinel-2 image time series acquired in 2016. Each image has a spatial resolution of 48 $\times$ 48 pixels and contains 13 channels. Every sample consists of a sequence of 30 time steps paired with a single segmentation-labeled image. We adopt the original division of the training, validation, and test sets.

In addition to the previously discussed SITS FMs, we also compare a fully supervised vision transformer network, TSViT \cite{tarasiou2023vits}, specifically developed for crop classification using longer time series. Table \ref{tab:mtlcc} presents the experimental results, which show that TSViT achieves higher accuracy than FMs such as SeCo, CACo, and GASSL. This improvement may be attributed to its attention mechanism that effectively encodes temporal information, whereas SeCo, CACo, and GASSL rely solely on a ResNet-50 encoder for extracting spatial context. With pre-training on MillionST, our TiMo significantly outperforms these models by achieving the highest accuracy, highlighting its strong capability to model long time series RS data.

\begin{table}
    \centering
    \caption{Comparison of different models on MTLCC dataset.}\setlength{\tabcolsep}{10pt}
    \label{tab:mtlcc}
    \resizebox{\linewidth}{!}{
    \begin{tabular}{lccc}
    \hline
        Method  &  Backbone &\#Parameter & mIoU   \\ 
        \hline
        TSViT\cite{tarasiou2023vits}&TSViT&2M&0.7617\\
  SeCo\cite{manas2021seasonal}&ResNet-50&26M&0.4309\\
    CACo\cite{mall2023change}&ResNet-50&26M&0.3766\\
GASSL\cite{ayush2021geography}&ResNet-50&26M&0.3798\\
          Prithvi-EO-2.0\cite{szwarcman2024prithvi}&ViT-Large&300M&0.6333\\
          SatMAE\cite{cong2022satmae}&ViT-Large&307M&0.6323 \\
          \hline
         TiMo-Base&TiMo-Base&91M&0.7655\\
         TiMo-Large&TiMo-Large&298M&\textbf{0.7718}\\
         \hline
    \end{tabular}
    }
\end{table}

\subsection{Flood Disaster Assessment}

Finally, we focus on flood detection. We begin by evaluating the methods using the Sen12Flood dataset \cite{rambour2020sen12}, which was designed for image classification. Sen12Flood includes both synthetic aperture radar (SAR) and multispectral images intended for multi-temporal flood detection, where images are classified as either \textit{flood} or \textit{non-flood}. The dataset comprises 336 sequences from regions in West and South-East Africa, the Middle East, and Australia. For our experiments, we use 30 Sentinel-2 optical (RGB) image sequences for training and the remainder for testing. On average, each sequence contains 14 temporal phases; for classification purposes, we randomly select three phases and use the corresponding RGB channels.

\begin{table}
\footnotesize
    \centering
    \caption{Comparison of different models on Sen12Flood dataset.}
    \label{tab:Sen12flood}
    \begin{tabular}{lcc}
    \hline
        Method & \#Parameter  & Acc(\%)\\
        \hline
        SeCo\cite{manas2021seasonal}&26M&72.35\\
        CACo\cite{mall2023change}&26M&75.52\\
        GASSL\cite{ayush2021geography}&26M&74.97\\
         SatMAE\cite{cong2022satmae}&300M& 81.86\\
         Prithvi-EO-2.0\cite{szwarcman2024prithvi}&307M& 80.11\\
         \hline
         TiMo-Base&91M &\textbf{86.89}\\
         TiMo-Large&298M &86.78 \\
         \hline
    \end{tabular}
\end{table}

Table~\ref{tab:Sen12flood} presents the results on the Sen12Flood dataset. Among all the methods compared, SatMAE demonstrates strong performance by slightly outperforming Prithvi-EO-2.0, while our proposed TiMo achieves the highest accuracy. Nevertheless, we observe that TiMo-Large performs only on par with the base version, possibly due to overfitting caused by the task's simplicity.

We then refine our analysis to locate the precise pixel-level regions affected by floods by framing the problem as a segmentation task. To further assess cross-domain performance, we utilize the KuroSiWo dataset~\cite{bountos2023kuro}, which consists of SAR images captured by the Sentinel-1 satellite across three temporal phases, with two channels per image. Each sample in KuroSiWo includes two pre-flood images and one post-flood image. The corresponding label is a mono-temporal, pixel-level annotation categorizing each pixel into one of three classes: \textit{No water}, \textit{Permanent waters}, and \textit{Floods}. In our experiments, we use the first fold of the dataset, comprising 7,103 time-series image samples, and split it into training, validation, and testing sets in a 3:1:1 ratio.  

The experimental results are presented in Table \ref{tab:kurosiwo}. As shown, TiMo-Base achieves competitive performance compared to other RSFMs. Notably, TiMo-Large significantly improves accuracy, achieving an impressive mIoU of 0.6711, despite the challenging transfer from optical to SAR images. These findings highlight the practicality and effectiveness of the proposed SITS FMs—TiMo.

\begin{table}
    \centering
    \caption{Comparison of different models on KuroSiwo dataset.}
    \label{tab:kurosiwo}
    \resizebox{\linewidth}{!}{
    \begin{tabular}{llccc}
    \hline
        Method  &  Backbone & \#Parameter & mIoU \\
        \hline
          SeCo\cite{manas2021seasonal}&ResNet-50&26M&0.5526 \\
          CACo\cite{mall2023change}&ResNet-50&26M&0.5223 \\
          GASSL\cite{ayush2021geography}&ResNet-50&26M&0.5742\\
         Prithvi-EO-2.0\cite{szwarcman2024prithvi}&ViT-Large&300M&0.6367\\
          SatMAE\cite{cong2022satmae}&ViT-Large&307M&0.4707\\
          \hline
         TiMo-Base&TiMo-Base&91M&0.6680 \\
         TiMo-Large&TiMo-Large&298M&\textbf{0.6711} \\
        
         \hline
    \end{tabular}
    }
\end{table}

\subsection{Ablation Study}

We conduct an ablation study to evaluate the impact of the model's structure and pre-training strategy. The results, presented in Table \ref{tab:ablationAtt}, examine the effect of different attention mechanisms. We compare six configurations, each defined by a unique combination of attention mechanisms. Our findings show that using MHSA in all blocks leads to high computational complexity and excessive GPU memory usage, causing the model to run out of memory. By replacing MHSA with the proposed STGA or D-STGA, we consistently achieve competitive performance compared to existing FMs (as shown in Table \ref{deforestTable}), validating that the proposed attention mechanisms are more effective at capturing spatiotemporal dynamics. Furthermore, we observe a significant reduction in computational complexity (FLOPs) with our attention mechanisms, particularly compared to the default MHSA used in vanilla vision transformers (\ie, M-M-M-M), demonstrating their higher efficiency. When comparing the second and third configurations, we find that D-M-M-M yields better accuracy, emphasizing the importance of modeling spatiotemporal relationships through the differential design. Additionally, FLOPs are further reduced, proving that this approach saves computational resources by minimizing the number of vector inner products. Finally, as we increase the number of blocks using D-STGA, we identify that the D-D-M-M configuration strikes the best balance between accuracy and efficiency.

\begin{table}
\footnotesize
\centering
\caption{
Ablation study on the MultiEarth dataset, where ``X'' in ``X-X-X-X'' denotes the attention mechanism used at each stage. S: STGA, D: D-STGA, M: MHSA, and OOM: Out of Memory. Memory was measured on an NVIDIA RTX 4090 GPU.
}\label{tab:ablationAtt}
\begin{tabular}{lccc} 
\hline
Method  & mIoU& FLOPs (G)& GPU Memory(M) \\ 

        \hline
        M-M-M-M  & - & 642.15 & OOM \\
        S-M-M-M  & 0.7827 &436.87& 15,284 \\
          D-M-M-M&0.7836 &402.52& 15,332 \\
          D-D-M-M &\textbf{0.7860}&372.60& 14,222 \\
          D-D-D-M&0.7827&339.10& 13129\\
          D-D-D-D& 0.7628 & 333.64 & 13,122 \\
          
\hline
\end{tabular}
\end{table}

\begin{table}
\small
\centering
\caption{Ablation study on the sampling strategy of pre-training, where the model is fine-tuning on the MultiEarth dataset.}
\begin{tabular}{lcc}
\hline
Method & Sampling & mIoU \\
        \hline
        TiMo-Base&Fixed&0.7839\\
         TiMo-Base&Random&\textbf{0.7860} \\
\hline
\end{tabular}
\label{random}
\end{table}

Since MillionST consists of 10 time phases, but our limited GPU memory only supports simultaneous processing of images with a temporal length of 3, we also compare different temporal sampling strategies during pre-training. Specifically, we evaluate the effects of using fixed versus random temporal sampling. In the random sampling approach, for each sample, 3 temporal images are randomly selected from the time series in every iteration. The results in Table \ref{random} show that random temporal sampling during pre-training leads to better performance compared to fixed sampling. This is likely because random sampling provides more diverse data, helping the model capture dynamic temporal variations in SITS, which in turn improves generalization and performance on downstream tasks.

\begin{figure}[t]
	
    \includegraphics[width=1\linewidth]{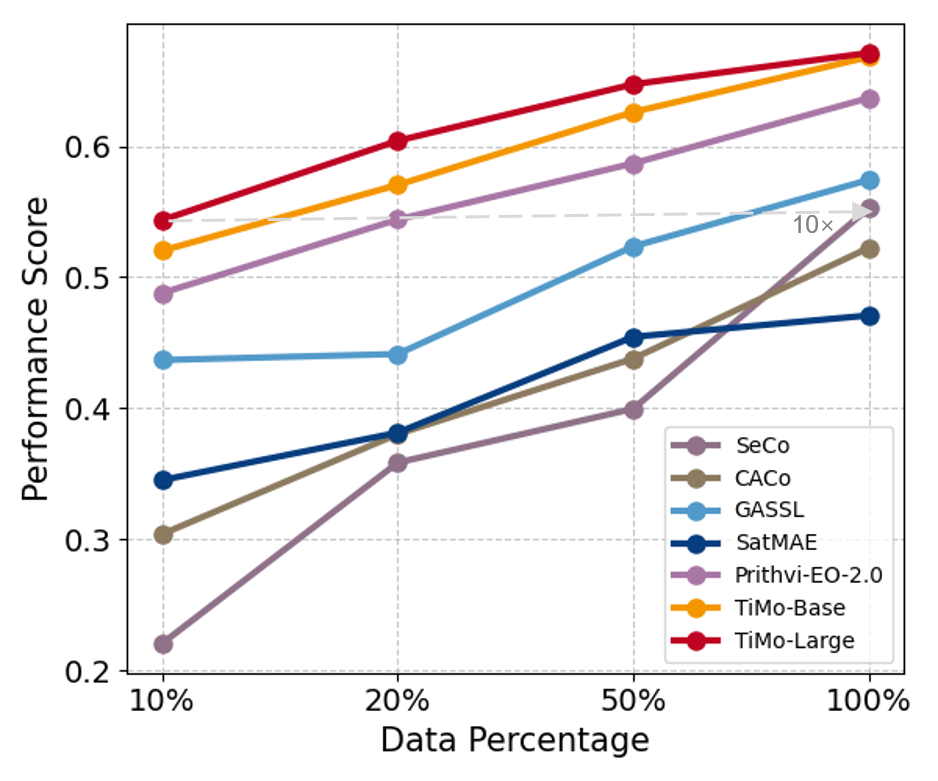}
	\caption{
    Experimental results of different SITS FMs trained with varying sample sizes on the KuroSiwo dataset.
    }
	\label{fewshot}
\end{figure}

\subsection{Data Efficiency of SITS FMs}

Data efficiency is a critical property of FMs \cite{zhang2023vitaev2,wang2022advancing,fm_sample_eff1,fm_sample_eff2}, especially when fine-tuning on downstream tasks with limited labeled data. To investigate this aspect in the context of TiMo, we conduct fine-tuning experiments on the KuroSiwo dataset~\cite{bountos2023kuro}, utilizing different proportions of the original training data—specifically 10\%, 20\%, and 50\%. We then evaluate the resulting models on the original test set. The experimental results, shown in Figure~\ref{fewshot}, demonstrate that our proposed methods consistently outperform other models across all training sample settings. 
Notably, TiMo-Large, trained on 50\% of the samples, outperforms Prithvi-EO-2.0, trained on the full dataset. With just 10\% of the data, TiMo-Large almost matches GASSL and SeCo and outperforms CaCo and SatMAE (all trained on the full dataset), highlighting its exceptional data efficiency.

\begin{figure}[t]
    \includegraphics[width=1\linewidth]{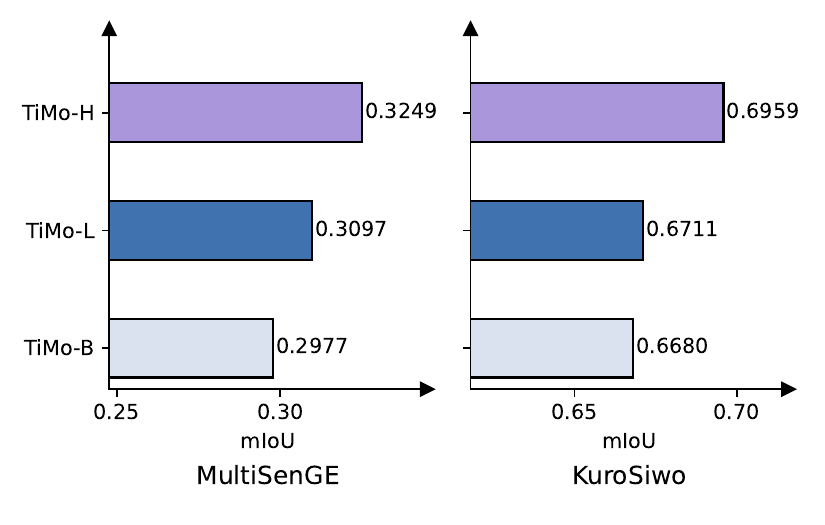}
	\caption{
    Fine-tuning performance of TiMo's variants on MultiSenGE and KuroSiwo datasets. TiMo-B, TiMo-L, and TiMo-H are short for TiMo-Base, TiMo-Large, and TiMo-Huge, respectively.}
	\label{scalability}
\end{figure}

\subsection{Scalability of TiMo}
To evaluate TiMo's scalability, we conduct fine-tuning experiments using TiMo's variants of different model sizes (91M, 298M, and 675M for TiMo-Base, TiMo-Large, and TiMo-Huge respectively) on two challenging datasets. As shown in Figure \ref{scalability}, larger TiMo backbones lead to better performance. Specifically, TiMo-Huge demonstrates a 9.14\% higher mIoU than TiMo-Base on MultiEarth dataset. Additionally, on the KuroSiwo dataset, TiMo-Huge exhibits a 4.18\% mIoU improvement over TiMo-Base. These results underscore TiMo's scalability across diverse tasks.

\section{Conclusion}

This study introduces TiMo, a novel hierarchical vision transformer foundation model explicitly designed for satellite image time series analysis. TiMo incorporates a novel spatiotemporal gyroscope attention (STGA) mechanism that effectively captures complex spatiotemporal relationships inherent in aligned SITS data, alongside an efficient D-STGA variant that leverages temporal differences to accelerate computation. To support effective pre-training, we curate MillionST, a large-scale dataset spanning 100,000 geographic locations with 10 temporal phases over five years, capturing diverse geospatial and seasonal variations. By adapting masked image modeling to this spatiotemporal context, TiMo learns generalizable representations for a wide range of downstream tasks. Extensive experiments across deforestation monitoring, land cover segmentation, and flood detection demonstrate TiMo's consistent superiority over existing models. Ablation studies further confirm the effectiveness of its design choices, highlighting its strengths in data efficiency and model scalability.
{
    \small
    \bibliographystyle{unsrt}
    \bibliography{main}
}
\clearpage
\appendix

\section*{Supplementary Material}
\section{Pre-training Dataset: MillionST}
To build the pre-training dataset, MillionST, we gathered Sentinel-2 multi-spectral Sequential Images Time Series (SITS) data from 100,000 sites across 1,317 cities in Europe, North Africa, and West Asia. These cities are among the world’s 10,000 most populous. Here, we provide a list of 100 cities among them.

[La Louvière, Seraing, Lierre, Binche, Gorna Oryahovitsa, Sandanski, Lida, Lugano, Biel/Bienne, Bellinzona, Köniz, Sion, Most, Frýdek-Místek, Mladá Boleslav, Přerov, Třinec, Znojmo, Příbram, Trutnov, Norderstedt, Garbsen, Langenhagen, Lingen, Elmshorn, Melle, Oberursel, Rodgau, Neustadt am Rübenberge, Lehrte, Falkensee, Dreieich, Wunstorf, Laatzen, Bensheim, Weißenfels, Buxtehude, Freital, Völklingen, Maintal, Ilmenau, Bitterfeld, Langen, Neu Isenburg, Papenburg, Königs Wusterhausen, Sankt Ingbert, Mörfelden-Walldorf, Seelze, Barsinghausen, Viernheim, Dietzenbach, Radebeul, Bad Vilbel, Wedel, Ahrensburg, Wernigerode, Lampertheim, Bad Nauheim, Hoyerswerda, Fürstenwalde, Achim, Georgsmarienhütte, Bramsche, Einbeck, Schönebeck, Burgdorf, Geesthacht, Riesa, Taunusstein, Andernach, Schwedt (Oder), Friedberg, Mīt Ghamr, Ziftá, Ra’s Ghārib, Vigo, Gijón, Badalona, Cartagena, Sabadell, Jerez de la Frontera, Mataró, Marbella, Algeciras, Lorca, El Puerto de Santa María, Mijas, Avilés, Rubí, Gandía, Benidorm, Benalmádena, Villanueva y Geltrú, La Línea de la Concepción, Arrecife, Granollers, Linares, Motril, Torrelavega]

\section{Implementation Details}
\subsection{Pre-training Details}
Following MAE \cite{he2022masked}, we use a mask ratio of 0.75 to mask partial tokens along both the spatial and temporal dimensions. We calculate the loss between the reconstructed and original pixels using normalized settings, as in MAE. Our experiments are carried out on NVIDIA RTX 4090 GPUs, with an effective batch size of 128, a base learning rate of 1.5$\times$$10^{-4}$, and a weight decay of 0.05. We pre-train the model for 100 epochs, including 10 warmup epochs, and we adopt the same learning rate scheduler and weight decay parameters as described in \cite{cong2022satmae}.

\subsection{Deforestation Monitoring}
We fine-tune TiMo and all comparative models on the MultiEarth dataset \cite{cha2023multiearth} for 30 epochs with a base learning rate of 0.0001 and a batch size of 2. For the decoder, we use UperNet to generate multi-temporal segmentation images for each timestamp. We fine-tune all models using four Nvidia RTX 4090 GPUs for 30 epochs with a base learning rate of 0.0001. We also employ the AdamW optimizer with a weight decay of 0.01.

\subsection{Multi-class Land Cover Segmentation}
 For fine-tuning on the MultiSenGE dataset \cite{wenger2022multisenge}, the input consists of multi-temporal images, while the output is a single land-cover segmentation map. Since the encoder generates multi-temporal features, we use UperNet to recover the spatial resolution of each image. We then apply a Max Pooling layer on the temporal axis to generate a mono-temporal feature.  We fine-tune all models using Nvidia RTX 4090 GPUs for 30 epochs with a base learning rate of 0.0001, a batch size of 2, and the AdamW optimizer with default settings. 

\subsection{Crop Type Classification}
Similar to the MultiSenGE dataset \cite{wenger2022multisenge}, the MTLCC dataset contains multi-temporal images and mono-temporal, pixel-level crop type annotations. We fine-tuned all models for 50 epochs using a base learning rate of 0.01, a batch size of 4, and the AdamW optimizer. The same decoder structure used for the MultiSenGE dataset was also employed for fine-tuning on the MTLCC dataset.

\subsection{Flood Disaster Assessment}

For the Sen12Flood dataset \cite{rambour2020sen12}, we train the models for 30 epochs with a base learning rate of 0.005 on the Sen12Flood dataset, using the AdamW optimizer with a weight decay of 0.05. We adopt a cosine decay learning rate schedule. Since this task aims to generate image-level prediction results, we apply average pooling to the spatial dimension, following the approach of \cite{he2022masked}. 

For the KuroSiwo dataset \cite{bountos2023kuro}, we train the models for 30 epochs with a base learning rate of 0.001. The same decoder structure used for the MultiSenGE dataset is also employed for fine-tuning on the KuroSiwo dataset.

\section{More Qualitative Results}

\begin{figure*}
\centering
    \hspace{-3em}
	\includegraphics[width=0.75\linewidth]{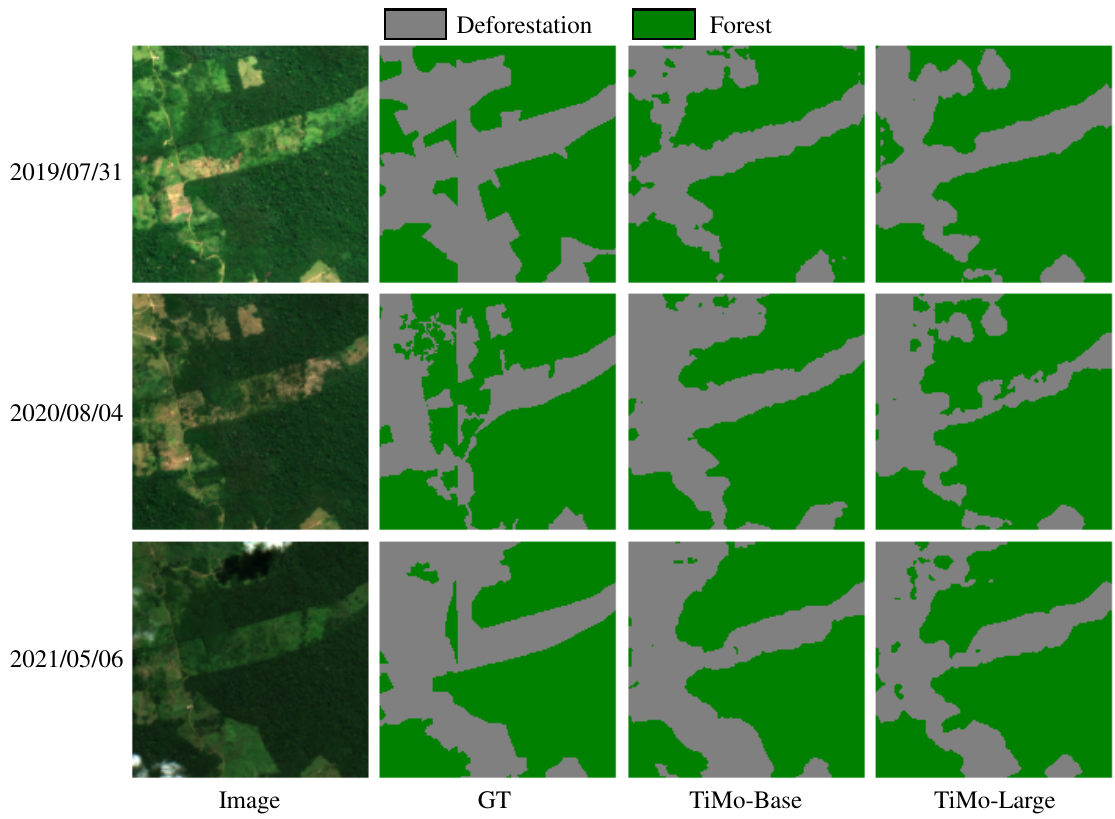}
	\caption{Visualization of the prediction results on MultiEarth dataset. GT denotes ground truth.}
	\label{MultiEarthvisual}
\end{figure*}

\begin{figure*}
\centering
	\includegraphics[width=0.75\linewidth]{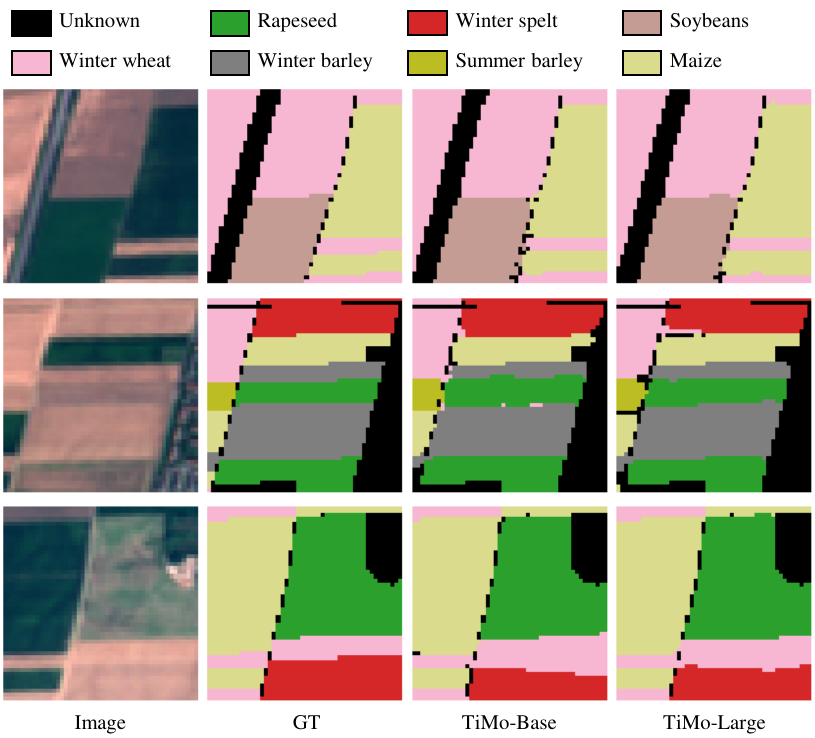}
	\caption{Visualization of the prediction results on MTLCC dataset. GT denotes ground truth.}
	\label{MTLCCvisual}
\end{figure*}

\begin{figure*}[t] 
\centering
	\includegraphics[width=0.8\linewidth]{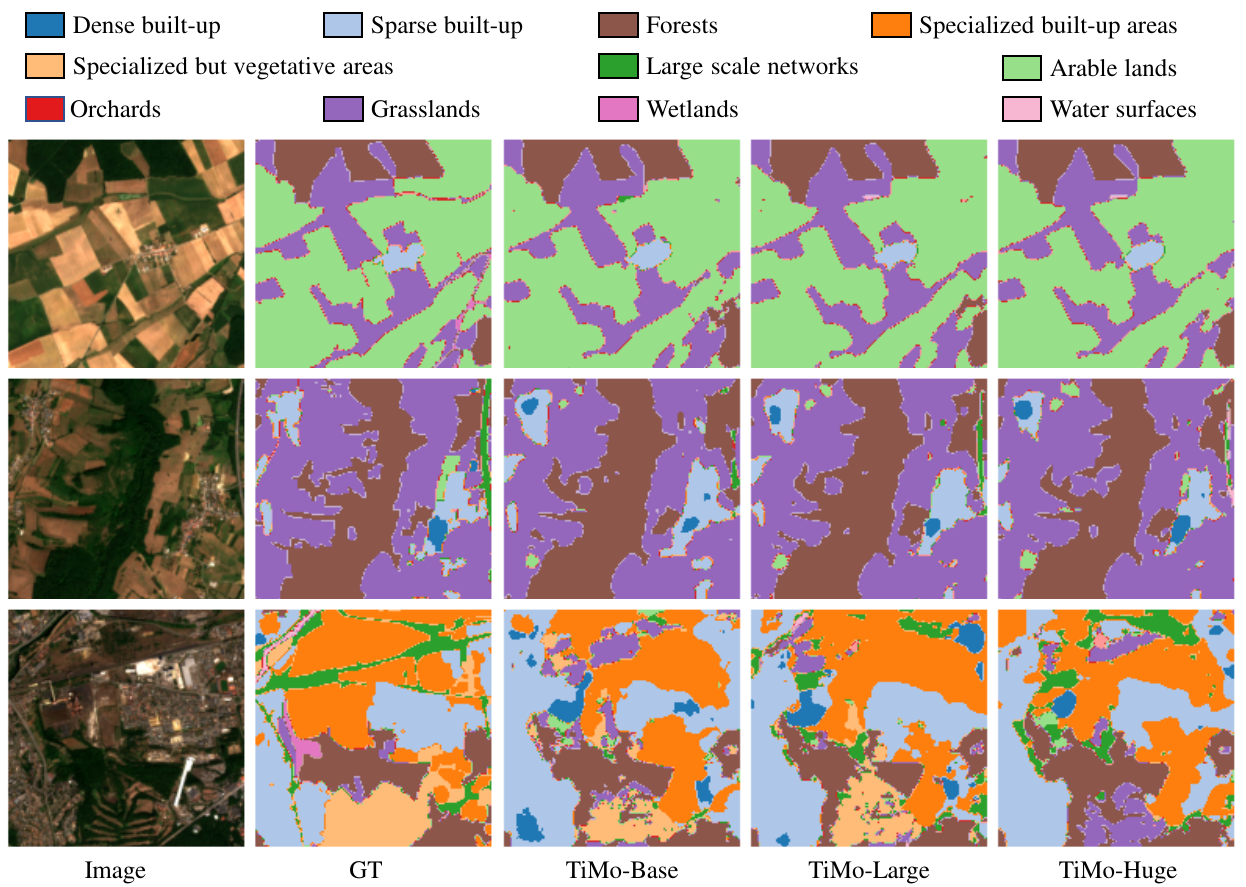}
	\caption{Visualization of the prediction results on MultiSenGE dataset. GT denotes ground truth.}
	\label{MultiSenGE}
\end{figure*}

\begin{figure*}
\centering
	\includegraphics[width=0.8\linewidth]{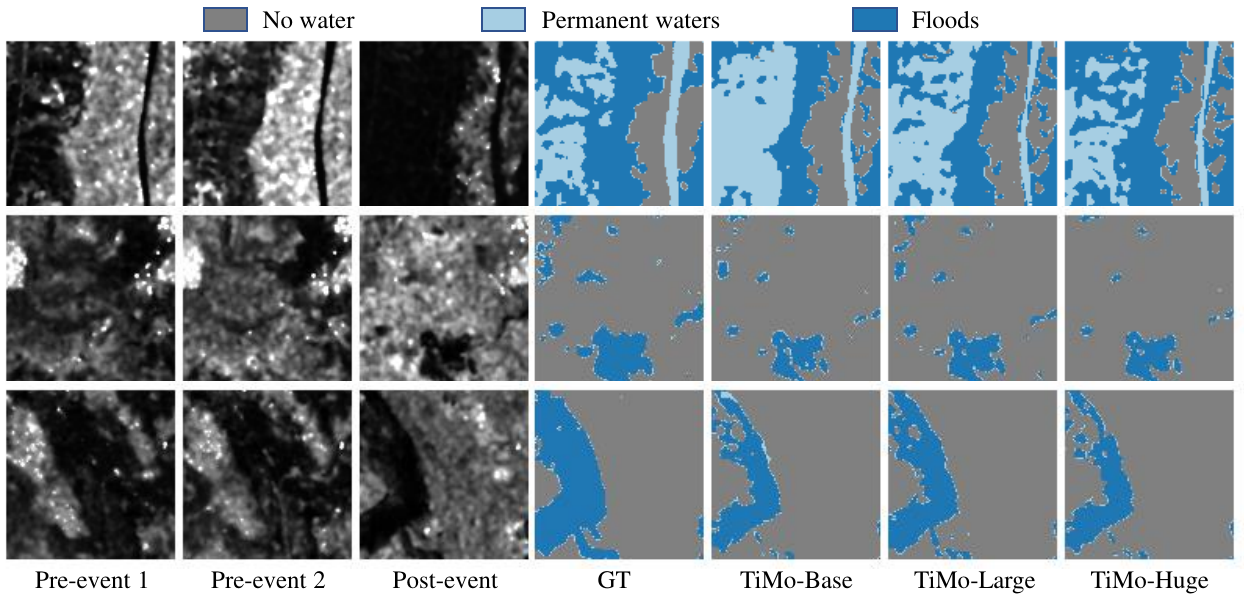}
	\caption{Visualization of the prediction results on KuroSiwo dataset. GT denotes ground truth.}
	\label{KuroSiwo}
\end{figure*}

Figures \ref{MultiEarthvisual}-\ref{KuroSiwo} present the prediction outcomes of our model across various applications, including deforestation monitoring, land cover segmentation, crop type classification, and flood disaster assessment. The results illustrate that our approach produces highly accurate segmentation maps from multi-temporal imagery, showcasing its exceptional ability to capture the spatiotemporal dynamics inherent in satellite image time series.

\section{Datasheet for MillionST}
\subsection{Motivation}

\noindent \textbf{1. For what purpose was the dataset created? Was there a specific task in mind? Was there a specific gap that needed to be filled? Please provide a description.}

\textbf{A1:} MillionST is designed to enable unsupervised pre-training of deep learning models using unlabeled satellite image time series (SITS) data. To address the limitations of many existing pre-training datasets, which often suffer from restricted temporal coverage, MillionST was created as a large-scale SITS dataset specifically tailored for spatiotemporal pre-training. The dataset comprises data collected from 100,000 geographic locations, with each location providing images from 10 timestamps spanning five years. Extensive spatiotemporal diversity in MillionST will facilitate the development of multi-temporal remote sensing community.

\noindent \textbf{2. Who created this dataset (e.g., which team, research group) and on behalf of which entity (e.g., company, institution, organization)?}

\textbf{A2:} This dataset is created by the authors of this paper.

\noindent \textbf{3. Who funded the creation of the dataset? If there is an associated grant, please provide the name of the grantor and the grant name and number.}

\textbf{A3:} N/A.

\subsection{Composition}

\noindent \textbf{1. What do the instances that comprise the dataset represent (e.g., documents, photos, people, countries)? Are there multiple types of instances(e.g., movies, users, and ratings; people and interactions between them; nodes and edges)? Please provide a description.}

\textbf{A1:} MillionST is comprised of image subsets captured by the Sentinel-2 satellite. Each subset consists of multispectral images from 10 timestamps and each image covers 2.65 $\times$ 2.65 km, consisting of 12 spectral bands at 10m, 20m, and 60m. 

\noindent \textbf{2. How many instances are there in total (of each type, if appropriate)?}

\textbf{A2:} MillionST has 100,000 instances, totally including 1,000,000 image patches.

\noindent \textbf{3. Does the dataset contain all possible instances or is it a sample (not necessarily random) of instances from a larger set? If the dataset is a sample, then what is the larger set? Is the sample representative of the larger set (e.g., geographic coverage)? If so, please describe how this representativeness was validated/verified. If it is not representative of the larger set, please describe why not (e.g., to cover a more diverse range of instances, because instances were withheld or unavailable).}

\textbf{A3:} MillionST is a sample of instances from a larger set. The larger set includes all Sentinel-2 images covering all timestamps around the world since the launch of the Sentinel-2 mission. Since the images in MillionST are mainly collected from Europe, North Africa, and West Asia, it cannot represent the larger set. Nevertheless, it contains images of 10 temporal phrases across five years, encompassing abundant spatiotemporal variations.

\noindent \textbf{4. What data does each instance consist of? “Raw” data (e.g., unprocessed text or images)or features? In either case, please provide a description.}

\textbf{A4:} Each instance consists of raw Sentinel-2 multispectral image patches from 10 temporal phrases.

\noindent \textbf{5. Is there a label or target associated with each instance? If so, please provide a description.}

\textbf{A5:} No, since this dataset is intended for spatiotemporal self-supervised learning.

\noindent \textbf{6. Is any information missing from individual instances? If so, please provide a description, explaining why this information is missing (e.g., because it was unavailable). This does not include intentionally removed information, but might include, e.g., redacted text.}

\textbf{A6:} No.

\noindent \textbf{7. Are relationships between individual instances made explicit (e.g., users’ movie ratings, social network links)? If so, please describe how these relationships are made explicit.}

\textbf{A7:} Yes. The relationships between different instances are shown in the folder name and \texttt{metadata.json}.

\noindent \textbf{8. Are there recommended data splits (e.g., training, development/validation, testing)? If so, please provide a description of these splits, explaining the rationale behind them.}

\textbf{A8:} Yes, we recommend utilizing the whole dataset for spatiotemporal self-supervised pre-training.

\noindent \textbf{9. Are there any errors, sources of noise, or redundancies in the dataset? If so, please provide a description.}

\textbf{A9:} No.

\noindent \textbf{10. Is the dataset self-contained, or does it link to or otherwise rely on external resources (e.g., websites, tweets, other datasets)? If it links to or relies on external resources, a) are there guarantees that they will exist, and remain constant, over time; b) are there official archival versions of the complete dataset (i.e., including the external resources as they existed at the time the dataset was created); c) are there any restrictions (e.g., licenses, fees) associated with any of the external resources that might apply to a future user? Please provide descriptions of all external resources and any restrictions associated with them, as well as links or other access points, as appropriate.}

\textbf{A10:} The dataset is self-contained since the samples are satellite imagery and can be downloaded from official websites. Users should refer to the official websites for information on any possible restrictions.

\noindent \textbf{11. Does the dataset contain data that might be considered confidential (e.g., data that is protected by legal privilege or by doctorpatient confidentiality, data that includes the content of individuals non-public communications)? If so, please provide a description.}

\textbf{A11:} No.

\noindent \textbf{12. Does the dataset contain data that, if viewed directly, might be offensive, insulting, threatening, or might otherwise cause anxiety? If so, please describe why.}

\textbf{A12:} No.

\subsection{Collection Process}

\noindent \textbf{1. How was the data associated with each instance acquired? Was the data directly observable (e.g., raw text, movie ratings), reported by subjects (e.g., survey responses), or indirectly inferred/derived from other data (e.g., part-of-speech tags, model-based guesses for age or language)? If data was reported by subjects or indirectly inferred/derived from other data, was the data validated/verified? If so, please describe how.}

\textbf{A1:} The data associated with each instance are directly observable, as they are stored in the GeoTIFF format and can be accessed via \href{https://rasterio.readthedocs.io/en/stable/}{Rasterio}.

\noindent \textbf{2. What mechanisms or procedures were used to collect the data (e.g., hardware apparatus or sensor, manual human curation, software program, software API)? How were these mechanisms or procedures validated?}

\textbf{A2:} The data after radiometric calibration and atmospheric correction are collected from Google Earth Engine through cloud removal, filtering of low cloud content images, temporal and spatial sampling, and cropping. All operations are controlled by Python scripts to operate the Google Earth Engine. The correctness of Python scripts is validated.

\noindent \textbf{3. If the dataset is a sample from a larger set, what was the sampling strategy (e.g., deterministic, probabilistic with specific sampling probabilities)?}

\textbf{A3:} The images in the dataset are probabilistically sampled.

Spatially, for each sample, the process begins with uniform sampling from 1,317 cities in Europe, North Africa, and West Asia. These cities are selected from the top 10,000 most populous cities globally. Subsequently, around each chosen city, Gaussian distribution sampling is performed, with the city center as the mean and a standard deviation of 50 kilometers.

Temporally, starting from a random time in the second half of 2017, timestamps are generated at six-month intervals until the first half of 2022, resulting in 10 timestamps. For each sample, a random offset is applied to these timestamps. Sampling is then conducted within a one-month window, covering 15 days before and after each offset timestamp. As a result, each location is associated with images corresponding to 10 timestamps.

\noindent \textbf{4. Who was involved in the data collection process (e.g., students, crowdworkers, contractors) and how were they compensated (e.g., how much were crowdworkers paid)?}

\textbf{A4:} The authors of this paper.

\noindent \textbf{5. Over what timeframe was the data collected? Does this timeframe match the creation timeframe of the data associated with the instances (e.g., recent crawl of old news articles)? If not, please describe the timeframe in which the data associated with the instances was created.}

\textbf{A5}: Since all operations are performed online, downloading is seriously affected by the network connection status, and collecting data costs about 1 month.

\subsection{Preprocessing/cleaning/labeling}

\noindent \textbf{1. Was any preprocessing/cleaning/labeling of the data done (e.g., discretization or bucketing, tokenization, part-of-speech tagging, SIFT feature extraction, removal of instances, processing of missing values)? If so, please provide a description. If not, you may skip the remainder of the questions in this section.}

\textbf{A1:} No.

\noindent \textbf{2. Was the “raw” data saved in addition to the preprocessed/cleaned/labeled data (e.g., to support unanticipated future uses)? If so, please provide a link or other access point to the “raw” data.}

\textbf{A2:} N/A.

\noindent \textbf{3. Is the software used to preprocess/clean/label the instances available? If so, please provide a link or other access point.}

\textbf{A3:} N/A.

\subsection{Uses}

\noindent \textbf{1. Has the dataset been used for any tasks already? If so, please provide a description.}

\textbf{A1:} No.

\noindent \textbf{2. Is there a repository that links to any or all papers or systems that use the dataset? If so, please provide a link or other access point.}

\textbf{A2:} N/A.

\noindent \textbf{3. What (other) tasks could the dataset be used for?}

\textbf{A3:} It can be used for the research of RS self-supervised learning, especially for spatiotemporal pre-training. 

\noindent \textbf{4. Is there anything about the composition of the dataset or the way it was collected and preprocessed/cleaned/labeled that might impact future uses? For example, is there anything that a future user might need to know to avoid uses that could result in unfair treatment of individuals or groups (e.g., stereotyping, quality of service issues) or other undesirable harms (e.g., financial harms, legal risks) If so, please provide a description. Is there anything a future user could do to mitigate these undesirable harms?}

\textbf{A4:} No.

\noindent \textbf{5. Are there tasks for which the dataset should not be used? If so, please provide a description.}

\textbf{A5:} No.

\subsection{Distribution}

\noindent \textbf{1. Will the dataset be distributed to third parties outside of the entity (e.g., company, institution, organization) on behalf of which the dataset was created? If so, please provide a description.}

\textbf{A1:} Yes. The dataset will be publicly available.

\noindent \textbf{2. How will the dataset will be distributed (e.g., tarball on website, API, GitHub)? Does the dataset have a digital object identifier (DOI)?}

\textbf{A2:} It will be publicly available on the project website.

\noindent \textbf{3. When will the dataset be distributed?}

\textbf{A3:} The dataset will be distributed once the paper is accepted after peer review.

\noindent \textbf{4. Will the dataset be distributed under a copyright or other intellectual property (IP) license, and/or under applicable terms of use (ToU)? If so, please describe this license and/or ToU, and provide a link or other access point to, or otherwise reproduce, any relevant licensing terms or ToU, as well as any fees associated with these restrictions.}

\textbf{A4:} It will be distributed under the \href{https://creativecommons.org/licenses/by-nc-sa/4.0/}{Creative Commons Attribution-NonCommercial-ShareAlike 4.0 License}.

\noindent \textbf{5. Have any third parties imposed IP-based or other restrictions on the data associated with the instances? If so, please describe these restrictions, and provide a link or other access point to, or otherwise reproduce, any relevant licensing terms, as well as any fees associated with these restrictions.}

\textbf{A5:} No.

\noindent \textbf{6. Do any export controls or other regulatory restrictions apply to the dataset or to individual instances? If so, please describe these restrictions, and provide a link or other access point to, or otherwise reproduce, any supporting documentation.}

\textbf{A6:} No.

\subsection{Maintenance}

\noindent \textbf{1. Who will be supporting/hosting/maintaining the dataset?}

\textbf{A1:} The authors.

\noindent \textbf{2. How can the owner/curator/manager of the dataset be contacted (e.g., email address)?}

\textbf{A2:} They can be contacted via email available on the project website.

\noindent \textbf{3. Is there an erratum? If so, please provide a link or other access point.}

\textbf{A3:} No. 

\noindent \textbf{4. Will the dataset be updated (e.g., to correct labeling errors, add new instances, delete instances)? If so, please describe how often, by whom, and how updates will be communicated to users (e.g., mailing list, GitHub)?}

\textbf{A4:} No.

\noindent \textbf{5. Will older versions of the dataset continue to be supported/hosted/maintained? If so, please describe how. If not, please describe how its obsolescence will be communicated to users.}

\textbf{A5:} N/A.

\noindent \textbf{6. If others want to extend/augment/build on/contribute to the dataset, is there a mechanism for them to do so? If so, please provide a description. Will these contributions be validated/verified? If so, please describe how. If not, why not? Is there a process for communicating/distributing these contributions to other users? If so, please provide a description.}

\textbf{A6:} N/A.



\end{document}